\documentclass{article}
\usepackage{ijcai18}

\usepackage{times}
\usepackage{xcolor}
\usepackage{soul}
\usepackage[utf8]{inputenc}
\usepackage[small]{caption}

\title{Differentiable Submodular Maximization }

\author{
	Sebastian Tschiatschek$^1$,
	Aytunc Sahin$^2$ \and
	Andreas Krause$^2$ 
	\\ 
   $^1$ Microsoft Research Cambridge\\
	$^2$ ETH Zurich \\
	sebastian.tschiatschek@microsoft.com,
	aytunc.sahin@inf.ethz.ch,
   krausea@ethz.ch
}

\usepackage{subcaption}
\usepackage{xspace}
\usepackage{booktabs}
\usepackage{etoolbox,siunitx}
\usepackage{amsmath,amsthm,amssymb}
\usepackage{mathtools}
\usepackage{algorithm,algorithmicx,algpseudocode}

\usepackage{titlesec}
\titlespacing*{\section}{0pt}{0.1\baselineskip}{0.1\baselineskip}

\robustify\bf

\newcommand{\FLIDD}{{FLID{\textsuperscript{D}}}}
\newcommand{\FLIDG}{{FLID{\textsuperscript{G}}}}

\newcommand{\genalg}{\textsc{ApproxMax}\xspace}

\newcommand{\alg}{\textsc{PD\textsuperscript{2}Greedy}\xspace}
\newcommand{\greedy}{\textsc{PGreedy}\xspace}

\newcommand{\pprobalg}{\ensuremath{P_\textsc{ApproxMax}}}
\newcommand{\groundset}{\ensuremath{\mathcal{V}}}

\newcommand{\pars}{\ensuremath{\theta}}
\newcommand{\gainadd}{\ensuremath{\Delta^+}}
\newcommand{\gainrem}{\ensuremath{\Delta^-}}

\newcommand{\D}{\ensuremath{\mathcal{D}}}
\newcommand{\Acc}{\textsc{Acc}\xspace}
\newcommand{\MRR}{\textsc{MRR}\xspace}
\newcommand{\llgain}{\textsc{rLL}\xspace}

\newcommand{\maximizer}{\ensuremath{X}}
\newcommand{\maximizers}{\ensuremath{\mathcal{X}}}

\newcommand{\opt}{\ensuremath{\textnormal{OPT}}}

\newcommand{\indicator}[1]{\ensuremath{\mathbf{1}_{#1}}}

\newtheorem{theorem}{Theorem}

\newcommand{\E}{\ensuremath{\mathbb{E}}}

\begin{document}

\maketitle

\begin{abstract}
We consider learning of submodular functions from data. These functions are important in machine learning and have a wide range of applications, e.g.\ data summarization, feature selection and active learning. Despite their combinatorial nature, submodular functions can be maximized approximately with strong theoretical guarantees in polynomial time. Typically, learning the submodular function and optimization of that function are treated separately, i.e.\ the function is first learned using a proxy objective and subsequently maximized. In contrast, we show how to perform learning and optimization jointly. By interpreting the output of greedy maximization algorithms as distributions over sequences of items and smoothening these distributions, we obtain a differentiable objective. In this way, we can differentiate through the maximization algorithms and optimize the model to work well with the optimization algorithm. We theoretically characterize the error made by our approach, yielding insights into the tradeoff of smoothness and accuracy. We demonstrate the effectiveness of our approach for jointly learning and optimizing on synthetic maximum cut data, and on real world applications such as product recommendation and image collection summarization.
\end{abstract}


\section{Introduction}

\noindent Many important applications require the prediction of set-valued outputs, e.g.\ product recommendation in which the output corresponds to a set of items that should be recommended to a user, object detection in images in which the output is a set of bounding boxes~\cite{felzenszwalb2010object}, or extractive summarization in which the output is a set of input objects (sentences in text summarization~\cite{lin2010multi}, images in image collection summarization~\cite{tschiatschek2014learning}, short scenes in video summarization~\cite{zhang2016video}). This problem is particularly challenging as the size of the output space (i.e.\ the number of possible subsets of some ground set, e.g.\ the set of all possible bounding boxes) increases exponentially with the size of the ground set.

The problem of learning to predict sets is most commonly approached by either specifying suitable set functions by hand or by first learning the set functions for the task at hand from data and then performing inference using the learned set function.
To enjoy strong theoretical guarantees for inference via maximization, we are interested in using set functions that are submodular.
The submodular set functions are fitted to the given training data (we consider the supervised case in which the training data contains the sets we wish to predict) using, for example, large margin methods~\cite{submodularshells,tschiatschek2014learning}, maximum likelihood estimation~\cite{gillenwater2014em,tschiatschek2016learning} or by minimizing some kind of set-valued loss function~\cite{dolhansky2016deep}. 
Finally, for inference, some variant of greedy algorithm is used for maximizing the learned set function.

This approach suffers from the problem that training and testing are performed differently which can lead to degraded performance. Furthermore, this approach circumvents end-to-end training of the considered functions. However, end-to-end training is desirable as it often leads to significant performance improvements as demonstrated on various domains recently~\cite{fasterrcnn,machinetranslation}.

We resolve this problem by proposing two differentiable algorithms for maximizing non-negative submodular set functions with cardinality constraints and without constraints. These algorithms can easily be integrated into existing deep learning architectures. The algorithms define a likelihood over sets and are used for \emph{both} training and testing. This enables end-to-end training of functions for predicting sets. Our algorithms are derived from the standard greedy algorithm~\cite{nemhauser1978analysis} and the double greedy algorithm~\cite{buchbinder2015tight} which was originally proposed for maximizing positive non-monotone submodular functions and provides a $\tfrac{1}{2}$ approximation guarantee. Interestingly, our algorithm for unconstrained maximization also provides approximation guarantees 
of the form $f(X_G) \geq \tfrac{1}{2} f(\textnormal{OPT})  - \epsilon(t)$, where $\epsilon(t)$ is a function of the parameter $t$ parameterizing our algorithm, $\textnormal{OPT}$ is an optimal solution and $X_G$ is the output of our algorithm.






Our main contributions are:
\begin{enumerate}
  \itemsep0em 
  \item We develop differentiable algorithms for end-to-end training of deep networks with set-valued outputs.
  \item We theoretically characterize the additional errors imposed by smoothing used in the derivation of our algorithm for unconstrained maximization.
  \item We demonstrate the excellent end-to-end learning performance of our approach on several challenging applications, including product recommendation and image collection summarization.
\end{enumerate}


\section{Notation \& Problem Setting}

{\bfseries Notation.} 
Let $\groundset = \{ e_1, e_2, \ldots \}$ be a ground set of items.
We want to perform subset selection from $\groundset$. In some parts of the paper we consider $\groundset$ to be fixed while in others $\groundset$ depends on the context. Note that we assume an arbitrary but fixed enumeration of the items in $\groundset$. Furthermore, let $f\colon 2^\groundset \rightarrow \mathbb{R}$ be a set function assigning a real-valued scalar to every set $S \subseteq \groundset$. The function $f$ is \emph{submodular} iff $f(A \cup \{e\}) - f(A) \geq f(B \cup \{e\}) - f(B)$ for all $e \in \groundset, A \subseteq B \subseteq \groundset \setminus \{e\}$. The function $f$ is \emph{monotone} if $f(A) \leq f(B)$ for all $A \subseteq B$ and \emph{non-monotone} otherwise. We denote the gain of adding an item $e \in \groundset$ to a set of items $S$ by $\gainadd_f(e \mid S) = f(S \cup \{e\}) - f(S)$ and the gain of removing the item $e \in S \subseteq \groundset$ from a set $S$ as $\gainrem_f(e \mid S) = f(S \setminus \{e\}) - f(S)$. We will omit the subscript $f$ whenever the function $f$ is clear from the context. For notational convenience we write $S+e$ instead of $S \cup \{e\}$ and $S-e$ instead of $S\setminus \{e\}$.


\noindent {\bfseries Problem setting.}
We consider the problem of learning an unknown submodular function $f(S \mid \groundset)$ from training data.
The training data is given in the form of a collection of maximizers $\maximizers$ (or approximate maximizers) of $f(S \mid \groundset)$ for different ground sets $\groundset$, i.e.\ $\maximizers=\{ (\groundset_1, \maximizer_1), \ldots, (\groundset_M, \maximizer_M)\}$.
We assume that the ground set is not only an abstract set of elements but that for every element in the ground set we have additional information, e.g.\ an associated vector of features.  
For instance, in the case of image collection summarization, the ground set $\groundset_i$ corresponds to the images (each element of $\groundset_i$ comes in form of an actual image in form of pixel values) in an image collection and the set $\maximizer_i$ is a human generated summary for that collection.
Note that the training data can contain multiple samples for the same ground set, e.g.\ in image collection summarization there can be multiple human generated summaries for some particular image collection.
The (approximate) maximizers in $\maximizers$ either maximize $f(S \mid \groundset)$  unconstrainedly, i.e.\ $\forall (\groundset, \maximizer) \in \maximizers$ it holds that $f(\maximizer \mid \groundset) \approx \max_{S \in \groundset} f(S)$ or cardinality constrainedly, i.e.\ $f(\maximizer) \approx \max_{S \in \groundset, |S|\leq k} f(S \mid \groundset)$ for some cardinality constraint $k$.
Given an algorithm \genalg that can approximately maximize a submodular function with or without cardinality constraints, our goal is to learn a set function $h_\pars(S \mid \groundset)$ parameterized by $\pars$ such that if it is maximized with \genalg, the returned solutions approximately maximize $f(S \mid \groundset)$.
For the image collection summarization application, maximizing $h_\pars(S \mid \groundset)$ with \genalg should generalize to new unseen ground sets $\groundset$, i.e.\ yield subsets of images that summarize an unseen image collection well, respectively.
For other applications, e.g.\ the product recommendation application, maximizing the function $h_\pars$ with \genalg should yield good solutions conditioned on the selection of a subset of items (see the experimental section for details).


\noindent {\bfseries Our approach---learning to greedily maximize.} 
We develop two probabilistic approximate maximization algorithms \genalg that take a function $h_\pars(\cdot | \groundset)$ parameterized by $\pars$ as input and output approximate maximizers of $h_\pars(\cdot | \groundset)$. These algorithms are based on the standard greedy algorithm and the double greedy algorithm and presented in detail in the next sections. 
Because of the probabilistic nature of the algorithms, they induce a distribution over sets $S \subseteq \groundset$ as $\pprobalg(S; h_\pars(\cdot|\groundset)) = P(\genalg(h_\pars(\cdot|\groundset)) = S)$.
We can thus equivalently view these probabilistic algorithms \genalg as distributions over sets.
To learn the parameters of the function $h_\pars$, we can thus maximize the likelihood $\maximizers$ under the induced distribution, i.e.\ we aim to find
\begin{equation}
  \pars^* = \arg\max_\pars \sum_{(\groundset,\maximizer) \in \maximizers} \log \pprobalg(\maximizer; h_\pars(\cdot | \groundset)). \label{eq:ml}
\end{equation}
The parameters $\pars^*$ according to the above equation intimately connect the function $h_{\pars^*}$, the algorithm \genalg, and the function $f$, i.e.\ \genalg computes approximate maximizers of $f$ when applied to $h_{\pars^*}$.
By ensuring that our developed approximate maximization algorithms have differentiable likelihoods, we can maximize our objective in~\eqref{eq:ml} using gradient based optimization techniques.
This enables the easy integration of our algorithms into deep learning architectures which are commonly optimized using stochastic gradient descent techniques.





\section{Differentiable Unconstrained Maximization}

\noindent In this section we present our probabilistic algorithm \alg for differentiable unconstrained maximization of non-negative submodular set functions. The algorithm is derived from the double greedy algorithm~\cite{buchbinder2015tight} and presented in Algorithm~\ref{alg:dgreedy}. 

The algorithm works by iterating through the items in $\groundset$ in a fixed order. In every iteration it computes the gain $a_i$ for adding the $i$th item to the set $X_{i-1}$ and the gain $b_i$ for removing the $i$th item from the set $Y_{i-1}$. It then compares these two gains and makes a probabilistic decision based on that comparison.

\begin{algorithm}
  \caption{\alg: Probabilistic diff.\ double-greedy}
  \label{alg:dgreedy}

  \begin{algorithmic}
    \Require Function $h_\pars\colon \groundset \rightarrow \mathbb{R}_{\geq 0}$ 
  
    \State $X_0 \leftarrow \emptyset$
    \State $Y_0 \leftarrow \groundset$
    \For {$i=1,\ldots,|\groundset|$}
      \State $a_i = h_\pars(X_{i-1} + e_i) - h_\pars(X_{i-1})$ \Comment $\hat{=}\gainadd(e_i \mid X_{i-1})$
      \State $b_i = h_\pars(Y_{i-1} - e_i) - h_\pars(Y_{i-1})$ \Comment $\hat{=}\gainrem(e_i \mid Y_{i-1})$
      \If {$g(a_i,b_i) \geq \mathcal{U}$} \Comment{$\mathcal{U}$ is uniform dist.\ on $[0,1]$}
        \State $X_i = X_{i-1} \cup \{e_i\}$
      \Else
        \State $Y_i = Y_{i-1} \setminus \{e_i\}$
      \EndIf
    \EndFor
    \State \Return{Approximate maximizer $X_{|\groundset|}$}
  \end{algorithmic}
\end{algorithm}

The deterministic and randomized version of the double greedy algorithm as presented in~~\cite{buchbinder2015tight} can be obtained from Algorithm~\ref{alg:dgreedy} by specific choices of the function $g(a_i,b_i)$. In particular, setting $g(a,b) = g_1(a,b) \coloneqq \indicator{a > b}$, where $\indicator{}$ is the indicator function, results in the deterministic double greedy algorithm with a $\tfrac{1}{3}$ approximation guarantee for maximizing positive non-monotone submodular functions. Setting $g(a,b) = g_2(a,b) \coloneqq \tfrac{[a]_+}{[a]_++[b]_+}$, where $[x]_+ = \max\{0,x\}$, results in the randomized double greedy algorithm with a $\tfrac{1}{2}$ approximation guarantee for maximization in expectation.

Note that the algorithm, independent of the particular choice of $g(a_i, b_i)$, induces a distribution over subsets of $\groundset$. To specify this distribution, let $x$ be a binary vector representing the set $X$ in form of a one-hot encoding for the assumed fixed ordering $e_1, e_2, \ldots$ of the elements in the ground set. Then, the distribution is
\begin{align}
  P_\alg(\maximizer) = \prod_{i=1}^{|\groundset|} &g(a_i, b_i)^{x_i} (1-g(a_i, b_i))^{1-x_i},  \label{eq:ll}
\end{align}
where $a_i = \gainadd( e_i \mid \{ e_j \mid j < i, x_j=1\})$ and $b_i=\gainrem(e_i \mid \{ e_j \in \groundset \mid j \geq i \textnormal{ or } x_j =1 \})$.
Obviously, this distribution depends on the order of the items which we assumed to be fixed. In practice, this order influences the maximization performance as substantiated by an example shortly. 

While Equation~\eqref{eq:ll} defines a distribution over sets it is hard to optimize for $g=g_1$ and $g=g_2$ because of the non-smoothness of these functions. Thus we propose the following two natural smooth alternatives for the function g:
\begin{itemize}
  \item $g(a,b) = g_3(a,b; t) \coloneqq \mu(a,b;t)$ is an approximation to the deterministic double greedy algorithm which becomes exact as $t \rightarrow 0$ where $\mu(a,b;t) = \tfrac{1}{1 + \exp(-(a-b)/t)}$
  \item $g(a,b) = g_4(a,b; t) \coloneqq \tfrac{a'}{a'+b'}$, where $a' = t \log(1+ \exp(\tfrac{a}{t}))$ and $b' =t \log(1+ \exp(\tfrac{b}{t}))$ is an approximation to the randomized double greedy algorithm which becomes exact as $t \rightarrow 0$.
\end{itemize}

Interestingly, choosing one of these smooth alternatives for the function $g(a,b)$ still results in an algorithm with 
approximation guarantees to the true optimum as substantiated in the following theorems.
\begin{theorem}
	\label{thm:dgreedy-rand}
	Let $\epsilon > 0$ and $g(a,b)=g_4(a,b;t)$. For $t <   \tfrac{2\epsilon}{|\groundset| \log(2)} $, the output $X_{|\groundset|}$ of Algorithm~\ref{alg:dgreedy} satisfies $f(X_{|\groundset|}) \geq \tfrac{1}{2}f(\textnormal{OPT}) - \epsilon$ in expectation, where $f(\textnormal{OPT})$ is an optimal solution.
\end{theorem}
Similarly, for the approximation to the deterministic algorithm we have the following theorem.
\begin{theorem}
	\label{thm:dgreedy-det}
	Let $\epsilon > 0$ and $g(a,b)=g_3(a,b;t)$. For $t <  \tfrac{3 \epsilon }{|\groundset| W(1/e)}$, the output $X_{|\groundset|}$ of Algorithm~\ref{alg:dgreedy} satisfies $f(X_{|\groundset|}) \geq \tfrac{1}{3}f(\textnormal{OPT}) - \epsilon$ in expectation, where $f(\textnormal{OPT})$ is an optimal solution.
\end{theorem}
The full proofs are omitted due to space constraints and available in the extended version of this paper~\cite{tschiatschek2018differentiable}. The idea of the proofs of Theorems~\ref{thm:dgreedy-rand} and~\ref{thm:dgreedy-det} is as follows: Instead of using $g_1$ and $g_2$ in Algorithm~\ref{alg:dgreedy}, we use $g_3$ and $g_4$, respectively. Note that as $t \rightarrow 0$, the approximation of $g_1$ by $g_3$ becomes exact (the same holds for the approximation of $g_2$ by $g_4$). For nonzero $t$, the considered approximations change the probabilities of selecting each item $e_i$ during the execution of Algorithm~\ref{alg:dgreedy} and introduce an (additive) error compared to executing Algorithm~\ref{alg:dgreedy} with $g_1$ and $g_2$, respectively. We can guarantee that this additive error is below any desired error $\epsilon/|\groundset|$ by choosing $t$ small enough. As the algorithm iterates over all items, the total error is then bounded by~$\epsilon$.

Theorems~\ref{thm:dgreedy-rand} and~\ref{thm:dgreedy-det} characterize a tradeoff between accuracy of the algorithm and smoothness of the approximation. High accuracy (small $\epsilon$) requires low smoothness (small $t$) and vice versa.

\noindent {\bfseries Speeding up computations.} 
Naively computing the likelihood in Equation~\eqref{eq:ll} requires $4 |\groundset|$ function evaluations. By keeping track of the function values $f(X_{i})$ and $f(Y_{i})$ this can immediately be reduced to $2 |\groundset|$ function evaluations. However, for large $|\groundset|$ this may still be prohibitive. Fortunately, many submodular functions allow for fast computation of all $a_i$ and $b_i$ values needed for evaluating~\eqref{eq:ll}. For example, for the modular function $f(S) = \sum_{e \in S} s_e$, where $s_e \in \mathbb{R}$, the gains for computing $\alg(\maximizer)$ are $a_i = s_{e_i}$ and $b_i = -s_{e_i}$. Another example is the facility location function $f(S) = \max_{e \in S} w_e$, where $w_e \in \mathbb{R}_{\geq 0}$. In that case, $a_i$ can be iteratively computed as $a_i = \max\{c_{i-1}, w_{e_i}\}$ where $c_{i} = \max\{c_{i-1}, w_{e_i}\}$ if $e_i \in X$ and $c_{i} = c_{i-1}$ otherwise (setting $c_0=-\infty$ for initialization).

\noindent{\bfseries Non non-negative functions.}
Some of the functions we are learning in the experiments in Section~\ref{sec:experiments} are not guaranteed to be non-negative over their whole domain.
This seems problematic because the above guarantees only hold for non-negative functions.
Note that any set function $\hat{f}(S)$ can be transformed into a non-negative set function $f(S)$ by defining $f(S) = \hat{f}(S) - \min_{S' \in \groundset} \hat{f}(S')$.
This transformation does not affect the gains $a_i$ and $b_i$ computed in Algorithm~\ref{alg:dgreedy}, however it changes the values of the maxima. 

\noindent {\bfseries Ordering of the items.}
The ordering of the items in the ground set influences the induced distributions over sets. Furthermore it can have a significant impact on the achieved maximization performance, although the theoretical guarantees are independent of this ordering.
To see this, consider the following toy example and \alg with $g=g_1$:
Let $\groundset=\{e_1, e_2\}$, $w_{e_1}=2,w_{e_2}=1$ and $f(S) = 2 + \max_{i \in S} w_i - |S|^2$.
This function is non-negative submodular with $f(\emptyset)=2, f(\{e_1\})=3, f(\{e_2\})=2, f(\groundset)=0$.
If the items in \alg were considered in the the order $(e_1, e_2)$, the algorithm would return $\{ e_2\}$ with a function value of $2$ while it would return $\{e_1\}$ with a function value of $3$ if the items were considered in the reversed order $(e_2,e_1)$.
Note that similar observations hold for all choices of $g$.



\noindent {\bfseries Connection to probabilistic submodular models.}
Algorithm \alg defines a distribution over sets and there is an interesting connection to probabilistic submodular models ~\cite{djolonga}
Specifically, for the case that $f(S)$ is a modular function, $g=g_3$ and $t=2$, the induced distribution of \alg corresponds to that of a log-modular distribution, i.e.\ items $e_i$ are contained in the output of \alg with probability $\sigma(f(\{e_i\})$.

\section{Diff.\ Cardinality Constrained Maximization}

In this section we present our algorithm for differentiable cardinality constrained maximization of submodular functions. We assume that the cardinality constraint is $k$, i.e.\ $\forall_{(\groundset,\maximizer) \in \maximizers}\colon f(\maximizer) \approx \max_{S \subseteq \groundset, |S|=k} f(S)$.

Our algorithm \greedy is presented in Algorithm~\ref{alg:greedy}. The algorithm iteratively builds up a solution of $k$ elements by adding one element in every iteration, starting from the empty set. The item added to the interim solution in iteration $i$ is selected randomly from all not yet selected items $\groundset \setminus X_{i-1}$, where the probability $p_{e \mid X_{i-1}}$ of selecting item $e$ depends on the gain of adding $e$ to $X_{i-1}$.
The selection probabilities are controlled by the temperature $t$. Assuming no ties, for $t \rightarrow 0$, \greedy is equivalent to the standard greedy algorithm, deterministically selecting the element with highest marginal gain in every iteration.

The algorithm naturally induces a differentiable distribution over sequences of items $\sigma=(e_1, \ldots, e_k)$ of length $k$ at temperature $t$ such that
\begin{align}
  P(\sigma) = \prod_{i=1}^{k} \frac{\exp(\tfrac{1}{t} \gainadd_{h_\pars}(\sigma_i \mid X_{i-1}))}{\sum_{e' \in \mathcal{V} \setminus X_{i-1}} \exp(\tfrac{1}{t} \gainadd_{h_\pars}(e' \mid X_{i-1}))},
\end{align}
where $X_{i-1} = \{\sigma_1, \ldots, \sigma_{i-1}\}$. From this distribution, we derive the probability of a set $S$ by summing over all sequences $\sigma$ of items consistent with $S$, i.e.
\begin{align}
  P(S) = \sum_{\sigma \in \Sigma(S)} P(\sigma), \label{eq:ll-greedy}
\end{align}
where $\Sigma(S)$ is the set of permutations of the elements in $S$. 
%

\begin{algorithm}
  \caption{\greedy: Probabilistic differentiable greedy}
  \label{alg:greedy}

  \begin{algorithmic}
    \Require Function $h_\pars\colon \groundset \rightarrow \mathbb{R}_{\geq 0}$, cardinality constraint $k$
  
    \State $X_0 \leftarrow \emptyset$
    \For {$i=1,\ldots,k$}
      \State $\mathcal{C} \leftarrow \groundset \setminus X_{i-1}$
      \State $\forall e \in \mathcal{C}\colon p_{e \mid X_{i-1}} \leftarrow \frac{\exp\big(\tfrac{1}{t} \gainadd_{h_\pars}(e \mid X_{i-1})\big)}{\sum_{e' \in \mathcal{C}} \exp\big(\tfrac{1}{t} \gainadd_{h_\pars}(e' \mid X_{i-1})\big)}$ 
      \State $e^* \leftarrow$ sample $e$ from $\mathcal{C}$ with probability $p_{e \mid X_{i-1}}$
      \State $X_i = X_{i-1} \cup \{e^*\}$
    \EndFor
    \State \Return{Approximate maximizer $X_{k}$}
  \end{algorithmic}
\end{algorithm}

As already briefly mentioned, the probability of a set $S$ depend on the temperature $t$.
For $t=0$ there is at most one permutation $\sigma$ of the items in $S$ for which $P(\sigma)$ is non-zero (assuming that there are no ties).
In contrast, for $t \rightarrow \infty$, $P(S \mid \sigma)$ is constant for all permutations. 
%
Similarly to the case of \alg, the choice of the temperature $t$ allows to trade-off accuracy (the standard greedy algorithm has a $(1-1/e)$ approximation guarantee for cardinality constraint maximization of non-negative monotone functions) and smoothness.
The higher the temperature, the smoother is the induced distribution but the \emph{further} are the made decisions from the greedy choice.

\noindent {\bfseries Learning with \greedy.} Given training data $\maximizers$, we can optimize the parameters of the function $h_\theta$ by maximizing the likelihood of $\maximizers$ under the distribution \eqref{eq:ll-greedy}. However, for large $k$ computing the summation over $\sigma \in \Sigma(S)$ is infeasible. We propose two approximations for this case:
\begin{itemize}
  \item If $t$ is \emph{small}, $P(S)$ can be accurately approximated by $P(\sigma^*(S))$ where $\sigma^*(S) = \arg \max_{\sigma \in \Sigma(S)} P(\sigma)$. However, to find $\sigma^*$ we would still have to search over all possible permutations of $S$. Hence, we propose to approximate $\sigma^*(S)$ by the permutation $\sigma^G(S)$ induced by the greedy order of the elements in $S$.
  \item Assuming there is no clear preference on the order of the items in $S$, i.e.\ $P(\sigma_i) \approx P(\sigma_j)$ for all $\sigma_i, \sigma_j \in \Sigma(S)$, we can approximate $P(S)$ as $P(S) \approx k! P(\sigma^R)$, where $\sigma^R$ is a random permutation of the items in $S$.
\end{itemize}

\noindent {\bfseries Testing.} If $k$ is known at test time, e.g.\ we aim to compute a summary of fixed size in the image collection summarization application, we  can simply execute the standard greedy algorithm using the learned function $h_\pars$. This returns an approximate MAP solution for $P(S)$. However, we can also generate approximate maximizers of $h_\pars$ by invoking Algorithm~\ref{alg:greedy}. This can be beneficial in cases where we want to propose several candidate summaries to a user to choose from.



\section{Experiments}
\label{sec:experiments}

\subsection{Maximum Cut} \label{maxcut}
Let $G(\groundset, E, w)$ be a weighted undirected graph, where $\groundset$ denotes the set of vertices, $E$ is the set of edges and $w$ denotes the set of non-negative edge weights $w_{ij}$, $\forall (i,j) \in E$. The maximum cut problem is to find a subset of nodes $S$ such that the cut value $f(S) = \sum_{i \in S} \sum_{j \in V \setminus S} w_{ij} $ is maximized. The cut function $f(S)$ is non-negative, non-monotone and submodular and has, for instance, been used in semi-supervised learning~\cite{maxcutsemisupervised}. 

We generate synthetic data for our maximum cut experiment as follows. We first generate $n$ vectors $x_1, \ldots, x_n \in \mathbb{R}^d$ by sampling their components uniformly from $[0,1]$  and arrange them in a $n \times d$ matrix $\mathbf{X}$. Using a projection matrix $\mathbf{P}$ of size $k \times d$, we project each $x_i$ to the lower dimensional space $\mathbb{R}^k$. The weights of the graph $w$ are defined using an RBF kernel $\mathbf{K}$ on these projected vectors. For the resulting graph instance $G(\groundset, E, w)$ we compute the maximum cut $S^{*} \subseteq V$ using IP (Integer Programming) formulation in Gurobi~\cite{gurobi}. 

We sample $\mathbf{X}_1, \mathbf{X}_2, \ldots, \mathbf{X}_m$ as described above and, using the $\textit{same}$ projection matrix $\mathbf{P}$ and RBF kernel $\mathbf{K}$, we generate weights $w^{i}$ for the $m$ graphs and find the maximum cut $S^{*}_{i}$ by IP for each of these graphs. From this data we compose our training set as $\mathcal{X}=\{ (\mathbf{X}_1, S^{*}_{1}), \ldots, (\mathbf{X}_m, S^{*}_{m})\}$. The test set is composed using the same strategy. 

In this experiment, our aim is to learn a projection matrix  $\hat{\mathbf{P}}$ such that \alg, if executed on the graph induced by the points $\hat{\mathbf{P}} \mathbf{X}_i$, returns $S^{*}_i$. Note that our goal is not necessarily to recover the original projection matrix $\mathbf{P}$---we are only interested in a projection matrix that allows us to find high value cuts through \alg.
For each element of the test set, we compute the corresponding graph using $\hat{\mathbf{P}}$ and find the maximum cut using  \alg and call the set as $S^{l}_{DG}$. We compute the corresponding graph using $\mathbf{P}$  and a random projection matrix, which acts as a baseline, $\mathbf{P}_{r}$ and find the maxcut using \alg; we refer to these cuts as $S^{o}_{DG}$ and $S^{r}_{DG}$  respectively. When the cut value is calculated using IP, we refer to it as $S^{o}_{IP}$. We compare the ratio of the each cut value calculated by \alg to the cut value calculated by IP.

Results are shown in Table~\ref{tab:maxcut}. We observe that cuts computed with \alg using the learned projection matrix outperform cuts computed via random projection matrices and the original projection matrix. The number of nodes in the graph is $n$, the temperature of the  \alg is $t$ and $0.3$ is used as a bandwidth parameter for the RBF kernel. The original points $x_i$ are in $\mathbb{R}^{10}$ and they are projected into $\mathbb{R}^5$ using the first 5 coordinates. Our training and test set consists of $800$ and $200$ elements, respectively. During training and testing, we used $g_4(a,b;t)$ to define the likelihood of sets as in Equation~\eqref{eq:ll}. For optimization, we used Adam~\cite{kingma2014adam} with batch size of $16$ and initial learning rate $0.02$. In every case, we benefit from learning the projection matrix.


\sisetup{detect-weight=true,detect-inline-weight=math}
\begin{table}[tb]

	\centering
	\small
\begin{tabular}{l*{3}{c}}
    \toprule
	Parameters          & $S^{l}_{DG}$  & $S^{o}_{DG}$& $S^{r}_{DG}$   \\
	\midrule
	n=20, t=$2^{-3}$ & \textbf{0.88 $\pm$ 0.008}	 & 0.86 $\pm$ 0.010 & 0.64	$\pm$ 0.012  \\
	n=20, t=$2^{-2}$   & \textbf{0.84 $\pm$ 0.007}	 & 0.80 $\pm$ 0.009 & 0.62	$\pm$ 0.012   \\
	n=20, t=$2^{-1}$           & \textbf{0.79		$\pm$ 0.006} & 0.74		$\pm$ 0.007 & 0.60	$\pm$ 0.008   \\
	n=20, t=1          & \textbf{0.74		$\pm$ 0.006} & 0.69		$\pm$ 0.007 & 0.60	$\pm$ 0.008   \\
	\bottomrule
\end{tabular}
	\caption{Performance of different cuts for varying temperature $t$ in \alg. The cuts $S^{l}_{DG}$ computed from \alg using the learned projection matrix outperforms cuts computed via random projection matrices and the original projection matrix. This indicates that the learned projection matrix is optimized to accommodate our probabilistic greedy algorithm.}
\label{tab:maxcut}
\end{table}


%
%



\subsection{On the Relation Between Learning and Temperature}

\begin{figure*}[th!]
	\centering
	\begin{subfigure}[t]{.5\linewidth}
	  \includegraphics{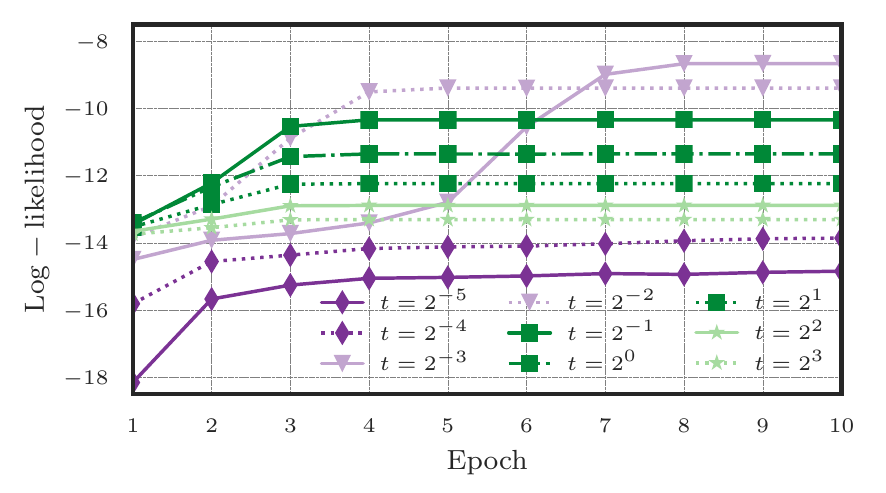}
   \end{subfigure}%
 	\begin{subfigure}[t]{.5\linewidth}
	  \includegraphics{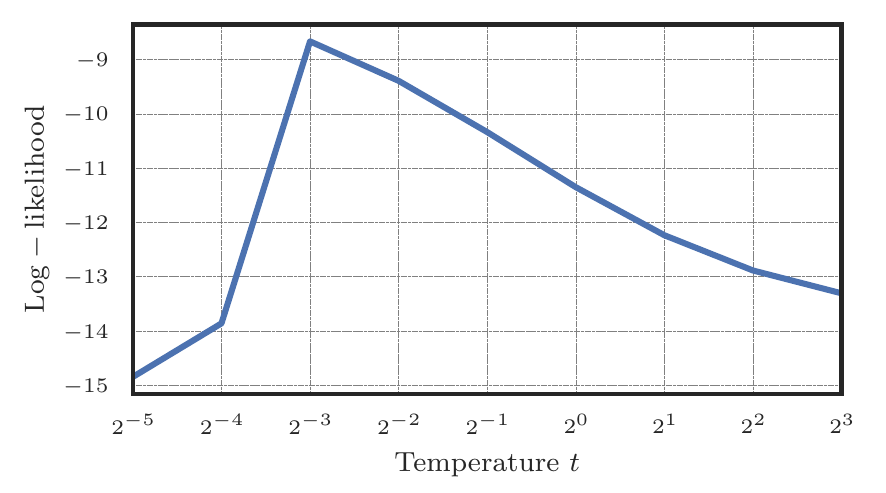}
   \end{subfigure}
	\caption{Relation between learning and temperature. (left) Training log-likelihoods over number of training epochs. Highest log-likelihoods are achieved for medium temperatures $t$ while low and high temperatures result in decreased log-likelihoods. (right) Log-likelihoods after convergence over temperature $t$. The highest log-likelihood is achieved for temperature $t=2^{-3}$.}
	\label{fig:figure1}
\end{figure*}

In this section we experimentally investigate the relation between temperature and the \emph{difficulty} of parameter learning, providing rather informal but intuitive explanations for the results of our paper.
Our findings are based on the experimental setting of Section~\ref{maxcut}.
Figure~\ref{fig:figure1} shows training log-likelihoods for different temperatures over training epochs on synthetic data.

For low temperatures $t$ ($t=2^{-5}$ and $t=2^{-4}$), we observe that the log-likelihood quickly converges to a (relatively) low log-likelihood level.
For these low temperatures, the link function $g_4$ is \emph{relatively close} to $g_2$, the standard link function of the randomized double greedy algorithm.
However, as $g_4$ becomes \emph{less smooth} with decreasing $t$, the corresponding gradients become non-informative and learning is more difficult. 

For high temperatures $t$ ($t=2^{2}$ and $t=2^{3}$), we also observe that the log-likelihoods converge to a (relatively) low log-likelihood level.
For these high temperatures, $g_4$ is \emph{very smooth} but \emph{far} from $g_2$. With increasing temperature, the probability distribution over sets becomes more uniform and non-informative. Because of the smoothness of $g_4$, the optimization becomes easier but it converges to a distribution that does not put much probability mass on the approximate maximizers that we are interested in.

For medium temperatures $t$ ($t=2^{-3}$ and $t=2^{-2}$) we observe the highest log-likelihoods.
This is also reflected in the test set performance in our other synthetic experiments, cf.~Table~\ref{tab:maxcut}.
For these medium temperatures, $g_4$ is \emph{smooth enough} to provide informative gradients and \emph{close enough} to the original link function $g_2$ to preserve the correct probability distribution over sets. 

To summarize, there is a trade-off between temperature and \emph{difficulty} of parameter learning.
Although Theorems~\ref{thm:dgreedy-rand} and~\ref{thm:dgreedy-det} suggest that using a temperatures $t$ as small as possible should lead to the smallest additive error compared to the corresponding variants of the double greedy algorithm, using low temperatures $t$ can make training \emph{difficult}.
Using high temperatures $t$ also results in bad log-likelihoods as the distribution over sets becomes uniform.
The plot on the right hand side of Figure~\ref{fig:figure1} shows the final log-likelihoods (at epoch 10) with different temperatures. From this plot, we can see that low and high temperatures end up in a lower log-likelihood than medium temperatures. 
Thus, it is advantageous to use temperatures $t$ that result in smooth link functions while simultaneously ensuring an informative probability distribution over sets. 

\subsection{Product Recommendation}

We consider the Amazon baby registry data~\cite{gillenwater2014em}.
 The data consists of baby registry data collected from Amazon and split into several datasets according 
to product categories, e.g.\ safety, strollers, etc. The datasets contain $5{,}500$ to $13{,}300$ registries of users. The data is processed using 10-fold cross-validation. As for \alg the order of items matters, we permute the items according to their empirical frequencies  for each dataset. 

For every dataset and fold, we fit a facility location diversity (FLID) type model~\cite{tschiatschek2016learning}, i.e.
\begin{equation*}
  f(S) = \sum_{i \in S} u_i + \underbrace{\sum_{d=1}^D \big( \max_{i \in S} w_{i,d} - \sum_{i \in S} w_{i,d}\big)}_{\textnormal{div}(S)},
\end{equation*}
where $u_i \in \mathbb{R}$ encodes the utility of product $i$ and $w_{i,d} \in \mathbb{R}_{\geq 0}$ encodes the $d$th latent property of item $i$. The intuitive idea behind the term $\textnormal{div}(S)$ is to encode repulsive dependencies between items that all have the $d$th latent property.

We train FLID models by optimizing the likelihood of the training data in Equation~\eqref{eq:ll} for $t=1$ and the sigmoid approximation (\FLIDD) and by optimizing the likelihood in Equation~\eqref{eq:ll-greedy} (\FLIDG) for $t=0.1$. For optimization we used Adam with a batch size of $1$ for \FLIDD{} and  with a batch size of $100$ for \FLIDG. We exhaustively computed the sum in Equation~\eqref{eq:ll-greedy} for sets with at most 5 items and otherwise approximated this sum by randomly selecting $5!$ of the summands (and scaling the result accordingly). We decayed the initial learning rate of $0.01$ with a decay factor of $0.9$ per epoch. We trained all models for 20 epochs.
The number of latent dimensions $D$ is chosen as 10 for ground sets with at most 40 items and 20 for larger ground sets.

\noindent {\bfseries Performance evaluation.} To assess the performance of our trained models we compute the following performance measures and compare them to the performance of \emph{modular} models (fully factorized models independently predicting the inclusion of any element in a set without considering correlations) and determinantal point process (DPPs) fitted by expectation maximization~\cite{gillenwater2014em}. Let $\D=\{ R_1, \ldots, R_N\}$ be a set of registries and let $\bar{\D}=\{ R \in \D \mid |R| > 1\}$ be the set of registries in $\D$ with at least 2 items.
\begin{itemize}
  \itemsep0em 
  \item \emph{Relative improvement in likelihood \llgain.} This measure quantifies the  increase in likelihood for FLID and DPPs over a modular model. The \llgain is computed as
    \begin{equation*}
      \llgain  = \frac{|\sum_{R \in \D} \log P(R) - \textnormal{LL}_\textnormal{mod}|}{\textnormal{LL}_\textnormal{mod}},
    \end{equation*}
    where $P(R)$ is the probability of $R$ for FLID/DPP and $\textnormal{LL}_\textnormal{mod} = \sum_{R \in \D} \log P_{\textnormal{modular}}(R) \log$ is the likelihood of the fully factorized model.
  
  \item \emph{Fill-in accuracy \Acc.} We test how accurately the considered models can predict items removed from a baby registry.  We compute the fill-in accuracy as
    \begin{equation*}
      \Acc(\D; \hat{f}) = \frac{1}{|\bar{\D}|} \sum_{R=\{ r_1, \ldots, r_k \} \in \bar{\D}} \sum_{i=1}^k \mathbf{1}_{\hat{f}(R - {r_i}) = r_i}
    \end{equation*}
    where $\hat{f}$ is the function used for prediction. For FLID and modular functions, $\hat{f}(S) = \arg \max_{e \in \groundset \setminus S} P(S + e)$. In the case of DPPs, $\hat{f}$ returns the element with largest marginal conditioned on $S$.
    
  \item \emph{Mean reciprocal rank \MRR.} Instead of only accounting correct predictions, this measure considers the order in which the models would predict items. Formally, \MRR is defined as
    \begin{equation*}
      \MRR(\D; \hat{f}) = \frac{1}{|\bar{\D}|} \sum_{R=\{ r_1, \ldots, r_k \} \in \bar{\D}} \sum_{i=1}^k \frac{1}{r(r_i;R - {r_i})},
    \end{equation*}  
    where $r(r_i; R-r_i)=k$ if the item $r_i$ would be predicted as the $k$th item. 

\end{itemize}

The results are shown in Table~\ref{tab:amazon}. We observe that FLID in most cases outperforms the modular model. The \llgain of DPPs is in general higher than that of \FLIDD.
In many cases \FLIDG{} significantly outperforms DPPs and \FLIDD{} in terms of fill-in accuracy and MRR. 


\sisetup{detect-weight=true,detect-inline-weight=math}
\begin{table*}[ht]

  \centering
  \begin{tabular}{lS[table-format=2.2]S[table-format=2.2]S[table-format=2.2]S[table-format=2.2]S[table-format=2.2]S[table-format=2.2]S[table-format=2.2]S[table-format=2.2]S[table-format=2.2]S[table-format=2.2]}
    \toprule
     & \multicolumn{2}{c}{\bfseries \llgain [\%]} & \multicolumn{4}{c}{\bfseries \Acc [\%]} & \multicolumn{4}{c}{\bfseries  \MRR [--]} \\\cmidrule(lr){2-3}\cmidrule(lr){4-7}\cmidrule(lr){8-11}
     \bfseries Dataset 	& {DPP}		& \FLIDD	& {modular}	& {DPP}		& \FLIDD	& \FLIDG	& {modular}	& {DPP}		& \FLIDD 	& \FLIDG \\\midrule
    safety 				& 11.60		& \bf 12.47	& 16.06		& 16.20		& 16.67 	& \bf 16.72 & 29.55		& 30.02		& 30.34 	& \bf 30.36 \\
    carseats 			& 8.77		& \bf 8.78	& 13.76		& 14.72		& 15.03		& \bf 16.18	& 28.75		& 30.14		& 30.56 	& \bf 31.33 \\
    strollers 			& 8.54		& \bf 9.58	& 14.42		& 16.79		& 19.69		& \bf 21.75	& 27.36		& 29.90		& 32.95 	& \bf 34.49\\
    furniture 			& 10.53		& \bf 10.59	& 15.82		& 16.08		& 15.93		& \bf 17.21	& 30.11		& 30.73		& 30.35 	& \bf 31.21 \\
    health 				& \bf 2.89	& 1.83		& 9.94		& 10.32		& 10.30		& \bf 12.07	& 21.42		& 22.31		& 21.87 	& \bf 23.26\\
    bath 				& \bf 2.66	& 1.65		& 7.30 		& 8.68	 	& 8.24		& \bf 9.90	& 16.06 	& 17.32		& 17.13 	& \bf 18.88\\
    media 				& \bf 2.35	& 1.19		& 9.31		& 10.07		& 10.04		& \bf 14.31	& 20.29		& 21.70		& 21.86 	& \bf 26.14 \\
    toys				& \bf 2.22	& 1.04		& 9.85		& 11.65		& 11.59		& \bf 14.78	& 21.93		& 23.40		& 23.64 	& \bf 26.23\\
    bedding 			& \bf 1.55	& 0.24		& 17.17 	& 17.43		& 16.87		& \bf 17.53	& 27.59		& \bf 28.04		& 27.59		& 28.01 \\
    apparel 			& \bf 0.93	& 0.00		& \bf 13.47	& 13.30		& 13.31		& \bf 13.28	& 20.88		& 21.19		& 21.16     & \bf 22.00\\
    diaper	 			& \bf 0.90	& 0.25		& 10.13		& 10.13		& 10.24		& \bf 14.55	& 18.03		& 18.56		& 19.10 	& \bf 23.78 \\
    gear 				& \bf 1.68	& 0.56		& 6.23		& 6.34		& 6.10		& \bf 6.46	& 14.80		& \bf 15.12	& 14.66 	& 14.71\\
    feeding 			& \bf 0.17	& 0.00		& 6.38		& 6.77		& 6.49		& \bf 9.24	& 14.53		& 15.22		& 15.02 	& \bf 18.00 \\
    \bottomrule
  \end{tabular}
 \caption{Performance of different models on the Amazon baby registry dataset. Performance of log-modular models fitted with maximum likelihood (modular), DPPs trained with the EM algorithm (DPP), FLID trained with \alg (\FLIDD), and FLID trained with \greedy (\FLIDG). 
	\FLIDD, \FLIDG and DPPs clearly outperform the modular model in most cases. \FLIDG outperforms \FLIDD and DPPs in many cases in terms of fill-in accuracy and MRR.
	\llgain is omitted for \FLIDG as it would require approximation due to the intractability of~\eqref{eq:ll-greedy} for large sets.
	A \llgain of $0.00$ indicates a negative likelihood improvement.}
\label{tab:amazon}
\end{table*}

\subsection{Image Collection Summarization}

We consider the problem of image collection summarization, i.e.\ the problem of selecting a subset of images $S$ of an image collection $\groundset$ such that the selected images summarize the content of the collection $\groundset$ well (e.g.\ $S$ covers all important scenes in $\groundset$ and the pictures in $S$ are diverse).

\noindent {\bfseries Data.} 
We use the image collection summarization data from~\cite{tschiatschek2014learning}. 
This data consists of 14 image collections $\groundset_1, \ldots, \groundset_{14}$, each  consisting of 100 images, i.e.\ $\groundset_i = \{ I^{(i)}_1, \ldots, I^{(i)}_{100} \}$.
The images capture diverse themes, e.g.\ traveling, shopping, etc.
For each image collection $i$, hundreds of human generated summaries were collected. We heuristically prune the original human summaries using the technique proposed in~\cite{tschiatschek2014learning}. The remaining human summaries $\mathcal{H}^{(i)} = \{H^{(i)}_1, \ldots, H^{(i)}_{N_i} \}$ are considered as approximate maximizers of an unknown function $f(S \mid \groundset_i)$ that we want to learn.

\noindent {\bfseries Evaluation, model, training. }
We follow the evaluation from~\cite{tschiatschek2014learning,dolhansky2016deep}.
That is, we train our model on 13 of the 14 image collections and test the model on the held out image collection.
We report the average VROUGE score~\cite{tschiatschek2014learning} on the hold out image collection.
The VROUGE score quantifies the quality of a summary for an image collection, capturing notions of coverage and diversity.

We train a model using the visual features computed in~\cite{tschiatschek2014learning}, similar as in~\cite{dolhansky2016deep}. 
Our model processes the 628 dimensional feature vector of each image $i$ by first compressing them into 100 dimensional vectors $h_i$ by a fully connected layer with ReLU activations. From this intermediate representation we compute a quality score $q_i$ for every image $i$ by another fully connected layer. From the intermediate representation we also compute means, maxes and variances across the images for every feature. These quantities are transformed into a threshold vector $b$ through another fully connected layer with ReLU activation ($b$ is 100 dimensional as the intermediate representation). The qualities, the threshold vector and intermediate representations parameterize a set function $f(S)=\sum_{i \in S} q_i + \sum_{j=1}^{100} \min(b_i, \sum_{i\in S} h_{i,j})$. We use dropout to avoid overfitting.

We optimize the expected VROUGE score induced by our proposed algorithm \greedy, using a temperature of $t=1$ and randomly sampled permutations of the items in the training samples (all human generated summaries consist of 10 images and hence computing the exact expected VROUGE score using~\eqref{eq:ll-greedy} is infeasible). For optimization, we used Adam~\cite{kingma2014adam} with a learning rate of $0.001$ and weight decay selected via cross-validation. 

\noindent {\bfseries Results.} 
We achieve an average normalized VROUGE score of 0.81. This is comparable to the performance achieved in~\cite{dolhansky2016deep}. The authors in~\cite{tschiatschek2014learning} achieved an average VROUGE score of 1.13 for their best model by learning a mixture of more than 500 hand specified component functions. In our experiments, we observed that our model easily overfits the training data. So more elaborate regularization methods could potentially improve the performance of our model further.

\section{Related Work}


While traditionally deep neural nets are used to learn parameters for a fixed size input vector, there is a growing literature on learning parameters of set functions with deep neural nets. For instance,~\cite{rezatofighi2017deepsetnet} considers learning the parameters of the likelihood of a set using deep neural nets and ~\cite{zaheer2017deep} designs specific layers to create permutation invariant and equivariant functions for set prediction.


\cite{learningsubmodular} considers learning submodular functions in a PAC-style framework and provides lower bounds on their learnability.   ~\cite{submodularshells,tschiatschek2014learning} learn mixtures of submodular functions by optimizing the weights using a large margin structured prediction framework. Since only the weights, not the components, are learned, the learning process is highly dependent on the representativeness of the functions used. ~\cite{dolhansky2016deep} develop a new class of submodular functions, similar to deep neural nets, and learn them in a maximum margin setting.

Recently, there has been some applications which combine deep learning and reinforcement learning to develop heuristics for combinatorial optimization problems. For instance,~\cite{maxcut} uses graph embeddings and reinforcement learning to learn heuristics for graph combinatorial optimization problems such as maximum cut.


\section{Conclusions}

We considered learning of submodular functions which are used for subset selection through greedy maximization. 
To this end, we proposed variants of two types of greedy algorithms that allow for approximate maximization of submodular set functions such that their output can be interpreted as a distribution over sets.
This distribution is differentiable, can be used within deep learning frameworks and enables gradient based learning.
Furthermore, we theoretically characterized the tradeoff of smoothness and accuracy of some of the considered algorithms.
We demonstrated the effectiveness of our approach on different applications, including max-cuts and product recommendation---observing good empirical performance.


\section*{Acknowledgements}

This research was partially supported by SNSF NRP 75 grant 407540 167212, SNSF CRSII2\_147633 and ERC StG 307036.

\clearpage

\bibliographystyle{named}
\bibliography{refs}

\clearpage

\appendix

\section{Proof of the Approximation Guarantee for Softplus Approximation  (Theorem~\ref{thm:dgreedy-rand})}
\begin{proof}
	  This proof closely follows the proof of Theorem $1.2$ in ~\cite{buchbinder2015tight}. 
	  We show below that for $t \leq \tfrac{\epsilon}{\log(2)}$ the following holds:
	    \begin{align}
	  &\E[f(\opt_{i-1}) - f(\opt_i)] \leq \label{eq:thm1_eq} \\
	  &\qquad \frac{1}{2}\E[f(X_i) - f(X_{i-1}) + f(Y_i) - f(Y_{i-1})] + \epsilon \nonumber
	  \end{align}
	  Summing over $i$ we get a telescopic sum. Collapsing it, we get
	  \begin{align*}
	  &\E[f(\opt_0) - f(\opt_n) ] = f(\opt) - \E[f(X_n)] \\
	  &\qquad \leq  \frac{1}{2} \E[f(X_n) - f(X_0) + f(Y_n) - f(Y_0)] + n \epsilon \\
	  &\qquad \leq \E[f(X_n)] + n\epsilon.
	  \end{align*}
	 Consequently, 
	   \begin{align*}
	 &\E[f(X_n)] \geq \frac{f(\opt)}{2} - \frac{n}{2} \epsilon. 
	 \end{align*}
	 We still need to prove inequality~\eqref{eq:thm1_eq}. We first bound the left hand side and rewrite the right hand side of~\eqref{eq:thm1_eq}:
	 \begin{enumerate}
	 	\item LHS:
	 	\begin{align*}
	 	\E[f(\opt_{i-1}) - f(\opt_i)] \leq \max\{p_a b, p_b a\},
	 	\end{align*}
	 	where $p_a = \frac{a'}{a' + b'}$ and $p_b = 1 - p_a$.
	 	\item RHS:
	 	\begin{align*}
	 &\frac{1}{2}	\E[f(X_i) - f(X_{i-1}) + f(Y_i) - f(Y_{i-1})] \nonumber \\
	 &\qquad = \frac{1}{2} (p_a a + p_b b).
	 	\end{align*}
	 \end{enumerate}
  Hence, to conclude with the statement of the theorem, it is sufficient to prove that $\max(\frac{a'b}{a' + b'},\frac{ab'}{a' + b'} ) \leq \frac{1}{2} (\frac{a'a}{a' + b'} + \frac{b' b}{a' + b'}) + \epsilon$ for $t \leq \tfrac{\epsilon}{\log(2)}$.
 W.l.o.g, assume $a'b \geq ab'$. Then there are the following cases:
 
   \begin{itemize}
 	\item {\bfseries Case 1. $a,b \geq 0$}. Let $\delta_a = a' - a$ and $\delta_b = b' - b$. Note that $\delta_a \leq t\log(2)$. Hence:
 	\begin{align*}
 	  &\epsilon \geq t \log(2) \geq t \log(2) \frac{b}{a'+ b'} \\
 	  &\Rightarrow \delta_a b \leq  \epsilon (a' + b') \\
 	  &\Rightarrow
 	(a + \delta_a)b - \frac{1}{2}(a + \delta_a)a - \frac{1}{2}(b + \delta_b)b \leq \epsilon (a' + b') \\
 	& \Leftrightarrow \max(\frac{a'b}{a' + b'},\frac{ab'}{a' + b'} ) \leq \frac{1}{2} (\frac{a'a}{a' + b'} + \frac{b' b}{a' + b'}) + \epsilon
 	\end{align*}
 	
 	\item {\bfseries Case 2. $a \leq 0, b \geq 0$}.
	Since $a \leq 0$, $a' > 0$ and $b \geq 0$, $b' \geq t \log(2)$. Thus, $a' + b' > t \log(2)$
	
	We can rewrite the statement that is sufficient to prove as $2a'b \leq a'a + b' b + \epsilon'$ with $\epsilon'= 2 \epsilon t \log(2) \leq 2\epsilon (a' + b')$.
	Equivalently:
	\begin{align*}
	  &0 \leq a'a - 2a'b + b'b + \epsilon' \\
	  \Leftrightarrow & 0 \leq (a'-b')^2 - (a'-b')^2 + a'a - 2a'b + b'b + \epsilon' \\
	  \Leftrightarrow & 0 \leq (a'-b')^2 + a'(a-a') + 2a'(b'-b)  \\
	                  &\qquad + b'(b-b') + \epsilon'
	\end{align*}
	The first and third term are non-negative. We show that $t \leq \tfrac{\epsilon}{\log(2)}$ is sufficient for the second and the fourth term to be not smaller than $-\epsilon'/2$.
	\begin{itemize}
	  \item For the second term we need $-\epsilon' / 2 \leq a'(a-a')$.
	  Observe that 
	    \begin{align*}
	      a-a' &= t \log(\exp(a/t)) - t \log(1 + \exp(a/t)) \\
	           &= -t  \log(1 + \exp(-a/t)).
	    \end{align*}
	    Hence we need $-\epsilon' / 2 \leq -a' t\log(1+\exp(-a/t))$.
	    Observe that $\log(1+\exp(a/t)) \log(1+\exp(-a/t))$ has maximum value $\log(2)^2$.
	    Thus it is sufficient to ensure that $-\epsilon' / 2 \leq -t^2 \log(2)^2$.
	    Note that $\epsilon'=  2 \epsilon t \log(2)$ and hence $t \leq \tfrac{\epsilon}{\log(2)}$ is sufficient. 
	    
	  \item The argument for the fourth term is analogeous to the one above, showing that $t \leq \tfrac{\epsilon}{\log(2)}$ is sufficient to prove our statement.
	  
	\end{itemize}
	It is hence sufficient to have $t \leq \tfrac{\epsilon}{\log(2)}$.
	
  \item {\bfseries Case 3. $a \geq 0, b \leq 0$}. This violates the assumption that $ a'b \geq ab'$.
  
  \item {\bfseries Case 4. $a \leq 0, b \leq 0$}. 
    Contradiction, since because of submodularity $a + b \geq 0$.
 	  
 \end{itemize}
\end{proof}


\section{Proof of the Approximation Guarantee for Sigmoid Approximation (Theorem~\ref{thm:dgreedy-det})}

\begin{proof}
	This proof closely follows the proof of Theorem $1.1$ in ~\cite{buchbinder2015tight}.
	
	We show below that for $t <  \tfrac{3 \epsilon }{|\groundset| W(1/e)}$ the following holds:
	\begin{align}
	&\E[f(\opt_{i-1}) - f(\opt_i)] \leq \label{eq:thm2_eq}\\
	&\qquad \E[f(X_i) - f(X_{i-1}) + f(Y_i) - f(Y_{i-1})] + \epsilon, \nonumber
	\end{align}
	From this it then follows that
	\begin{align*}
	&\E[f(\opt_0) - f(\opt_n) ] = f(\opt) - \E[f(X_n)] \\
	&\qquad \leq \E[f(X_n) - f(X_0) + f(Y_n) - f(Y_0)] + n \epsilon \\
	&\qquad \leq 2\E[f(X_n)] + n\epsilon.
	\end{align*}
	Consequently,
	\begin{align*}
	&\E[f(X_n)] \geq \frac{f(\opt)}{3} - \frac{n}{3} \epsilon.
	\end{align*}
	
	To prove the inequality~\eqref{eq:thm2_eq}, observe that:
	\begin{enumerate}
		\item LHS:
		\begin{align*}
		\E[f(\opt_{i-1}) - f(\opt_i)] \leq \max\{p_a b, p_b a\},
		\end{align*}
		where $p_a = \sigma(a-b)$ and $p_b = 1 - p_a$.
		\item RHS:
		\begin{align*}
		\E[f(X_n) - f(X_0) + f(Y_n) - f(Y_0)] = p_a a + p_b b.
		\end{align*}
	\end{enumerate}
	
	We consider three cases to prove $\max\{p_a b, p_b a\} \leq p_a a + p_b b + \epsilon$ which 
	is equivalent to $\max\{\sigma(a-b) b, \sigma(b-a) a\} \leq \sigma(a-b) a + \sigma(b-a) b + \epsilon$:
	\begin{itemize}
		\item {\bfseries Case 1. $a,b \geq 0$}. 
		If $a \geq b$, $\sigma(a-b) \geq 0.5$ and $a-b \geq 0$. Thus,
		\begin{align*}
		0 &= 0.5 (a-b) + 0.5 (b-a) \\
		& \leq \sigma(a-b)(a - b) + \sigma(b-a) (b-a).
		\end{align*}
		This implies
		\begin{align*}
		& \sigma(a-b) a + \sigma(b-a) b \geq p_a b + p_a b \geq \max\{p_a b, p_b a\}.
		\end{align*}
		For $a \leq b$ the same argumentation holds.
		
		\item {\bfseries Case 2. $a \geq 0, b \leq 0$}. In this case $\max\{p_a b, p_b a\} = p_b a$ and $\sigma(a-b) \geq 0.5$. Hence we need find $\epsilon$ such that
		\begin{align}
		0 \leq \sigma(a-b)a + \underbrace{\sigma(b-a)(b-a)}_{g(a,b)} + \epsilon
		\end{align}
		is satisfied to prove the above inequality. Note that the first term can be minimized by setting $a=0$ (which is always possible as the second term only depends on the difference $b-a$ and the third term is independent of both $a$ and $b$). Minimizing the term $g(a,b)$, which can be written as $g(x) = -\frac{1}{(1+e^{x/t})}x$ with $x=a-b$ yields that the minimum is achieved at $x^*=t W(1/e)+ t$ and evaluates to $g(x^*) = -t W(1/e)$, where $W$ is the Lambert-W function. Thus, having $t \leq \tfrac{1}{W(1/e)}  \epsilon$ ensures that the desired inequality holds.\footnote{Note that in fact we would want to solve an even more constrained problem ($a+b \geq 0$), which clearly also holds by the above selection of $t$.}
		
		\item {\bfseries Case 3. $a \leq 0, b \geq 0$}. Analogous to above. 
	\end{itemize}

\end{proof}

\end{document}